\documentclass[letterpaper, 10 pt, conference]{ieeeconf}  

\IEEEoverridecommandlockouts                              
\usepackage[utf8]{inputenc}
\usepackage[T1]{fontenc}

\usepackage[]{graphicx}
\usepackage{amsmath}
\usepackage{multirow}
\usepackage{blindtext}
\usepackage{subcaption}
\usepackage{xcolor}
\usepackage[export]{adjustbox}
\usepackage[normalem]{ulem}
\usepackage{float}
\usepackage{amssymb}
\usepackage{cite}
\usepackage{capt-of}
\usepackage{tabularx}
\usepackage{tikz}

\title{\LARGE \bf
Learning-Based Depth and Pose Estimation for Monocular Endoscope\\with Loss Generalization
}

\author{Aji Resindra Widya$^{1}$, Yusuke Monno$^{1}$, Masatoshi Okutomi$^{1}$,\\ Sho Suzuki$^{2}$, Takuji Gotoda$^{2}$, and Kenji Miki$^{3}$ 
\thanks{This work was supported by JSPS KAKENHI Grant Number 17H00744 and 21K12737.}
\thanks{$^{1}$A. R. Widya, Y. Monno, and M. Okutomi are with the Department of Systems and Control Engineering, School of Engineering, Tokyo Institute of Technology, Meguro-ku, Tokyo 152-8550, Japan
(e-mail: aresindra@ok.sc.e.titech.ac.jp; ymonno@ok.sc.e.titech.ac.jp; mxo@sc.e.titech.ac.jp).}
\thanks{$^{2}$S. Suzuki and T. Gotoda are with the Division of Gastroenterology and Hepatology, Department of Medicine, Nihon University School of Medicine, Chiyoda-ku, Tokyo 101-8309, Japan.}
\thanks{$^{3}$K. Miki is with the Department of Internal Medicine, Tsujinaka Hospital Kashiwanoha, Kashiwa-city, Chiba 277-0871, Japan.}%
}

\begin{document}

\maketitle
\thispagestyle{empty}
\pagestyle{empty}

\begin{abstract}

Gastroendoscopy has been a clinical standard for diagnosing and treating conditions that affect a part of a patient's digestive system, such as the stomach. Despite the fact that gastroendoscopy has a lot of advantages for patients, there exist some challenges for practitioners, such as the lack of 3D perception, including the depth and the endoscope pose information. Such challenges make navigating the endoscope and localizing any found lesion in a digestive tract difficult. To tackle these problems, deep learning-based approaches have been proposed to provide monocular gastroendoscopy with additional yet important depth and pose information. 
In this paper, we propose a novel supervised approach to train depth and pose estimation networks using consecutive endoscopy images to assist the endoscope navigation in the stomach. We firstly generate real depth and pose training data using our previously proposed whole stomach 3D reconstruction pipeline to avoid poor generalization ability between computer-generated (CG) models and real data for the stomach. In addition, we propose a novel generalized photometric loss function to avoid the complicated process of finding proper weights for balancing the depth and the pose loss terms, which is required for existing direct depth and pose supervision approaches. We then experimentally show that our proposed generalized loss performs better than existing direct supervision losses.

\end{abstract}

\section{Introduction}
Gastroendoscopy is one of the golden standards for finding and treating abnormalities inside a patient's digestive tract, including the stomach. Even though gastroendoscopy gives enormous advantages for the patient, such as no need for invasive surgeries, it is known that there exist some challenges for medical practitioners, such as the loss of depth perception and the difficulty in assessing the endoscope pose. These challenges lead to difficulties in navigating and understanding the scene captured by the endoscope system, making the localization of a found lesion hard for the practitioners.

Previous studies have proposed to reconstruct the 3D model of a whole stomach with its texture~\cite{widya2021stomach,widya2019whole,widya2021stomachIC} to provide a global view of the stomach and the estimated endoscope trajectory. It enables medical practitioners to perform a second inspection with more degree of freedom after an initial gastroendoscopy procedure. While the whole stomach 3D reconstruction provides the depth and the endoscope trajectory, the methods~\cite{widya2021stomach,widya2019whole} cannot be done alongside the gastroendoscopy procedure in real-time. 

Recent developments in endoscopy systems introduce a stereo camera to provide real-time depth information~\cite{maier_optical_2013,geng2014review,yang2016compact,Mahmoud2019live}. While the stereo endoscope solves the lack of depth perception, 
a monocular endoscope is still the mainstream system in clinical practice. As an alternative to the stereo endoscope, deep learning-based approaches have been proposed to provide depth information for monocular endoscopy~\cite{rau2019implicit,mahmood2018deep,widya2021self}.

To effectively tackle the endoscope navigation and the lesion localization challenges, only providing depth information is not enough. Both continuous depth and pose information are needed to address these challenges appropriately. Both supervised and self-supervised deep-learning-based approaches are heavily adopted to address simultaneous depth and pose estimation~\cite{turan2018deep,turan2018unsupervised,chen2019slam,ma2019real,ozyoruk2020endoslam}.
A commonly used supervision approach is to take the direct Euclidean distance losses for the predicted depth and pose in comparison with the respective ground truths or references~\cite{turan2018deep, chen2019slam}. In this approach, computer-generated (CG) and/or phantom models are commonly used for the training of depth and pose estimation networks, affecting the network generalization between CG and real data. In addition, the direct supervision approach needs balancing weights for depth and pose loss terms, which are difficult to search~\cite{kendall2015posenet}.

As a self-supervised approach, the study~\cite{turan2018unsupervised} uses consecutive frames as the inputs to train the network to simultaneously predict depth and pose by minimizing the photometric error of a view synthesis problem (image warping between consecutive frames based on the predicted depth and pose). 
Using the same principle, the study~\cite{ma2019real} trains a recurrent neural network to predict the depth and the pose and uses them as the inputs for standard SLAM~\cite{engel2017direct} for further refinement. Since it needs an additional hand-crafted method bootstrapped to the network architecture, this approach is not trainable in an end-to-end manner.
Even though a self-supervised training approach does not need labeled data for training and thus it is generally more convenient to train, its performance is yet to beat the supervised approach.

\begin{figure*}
    \captionsetup[subfigure]{justification=centering}
    \centering
    \begin{subfigure}{0.26\textwidth}
    \centering
    \includegraphics[width=1.045\columnwidth]{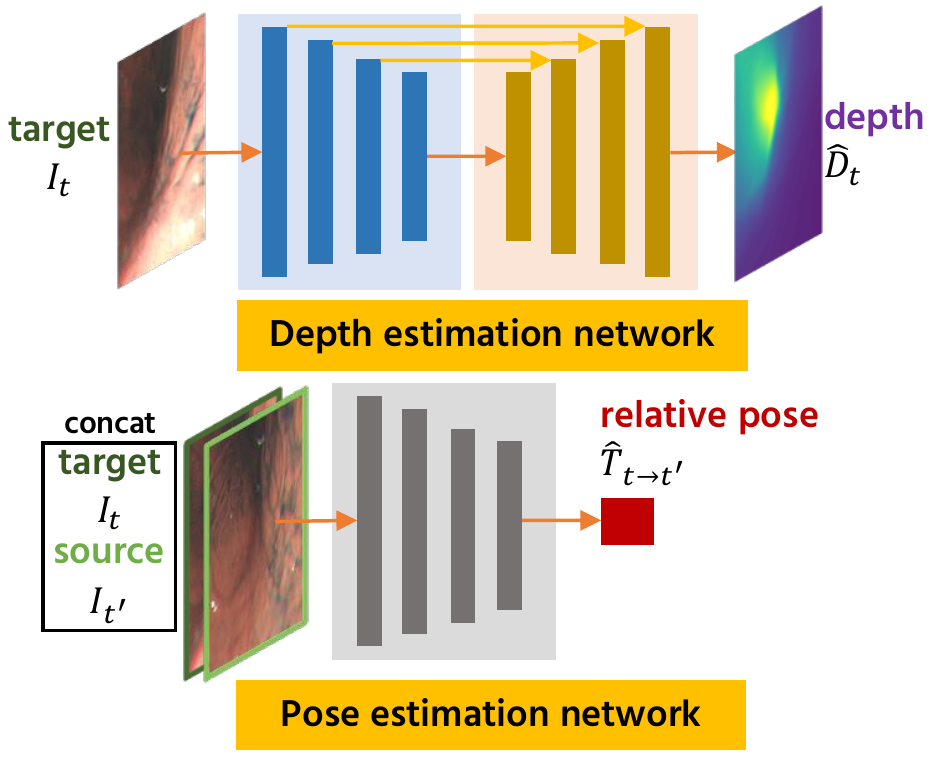}
    \caption{Network structure~\cite{godard2019digging}.}
    \end{subfigure}
    \begin{subfigure}{0.22\textwidth}
    \centering
    \vspace{2mm}
    \includegraphics[trim=25 25 25 25,clip,width=\columnwidth]{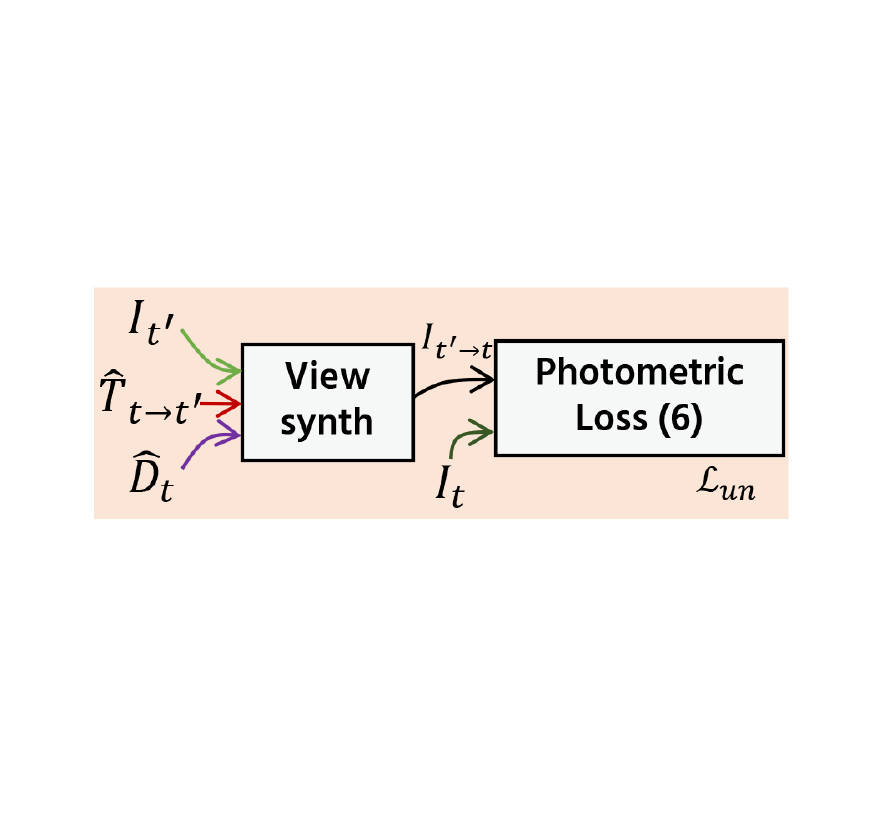}
    \vspace{2mm}
    \caption{Self-supervised photometric loss~\cite{godard2019digging}.}
    \end{subfigure}
    \begin{subfigure}{0.24\textwidth}
    \centering
    \vspace{5mm}
    \includegraphics[trim=0 0 0 0,clip,width=\columnwidth]{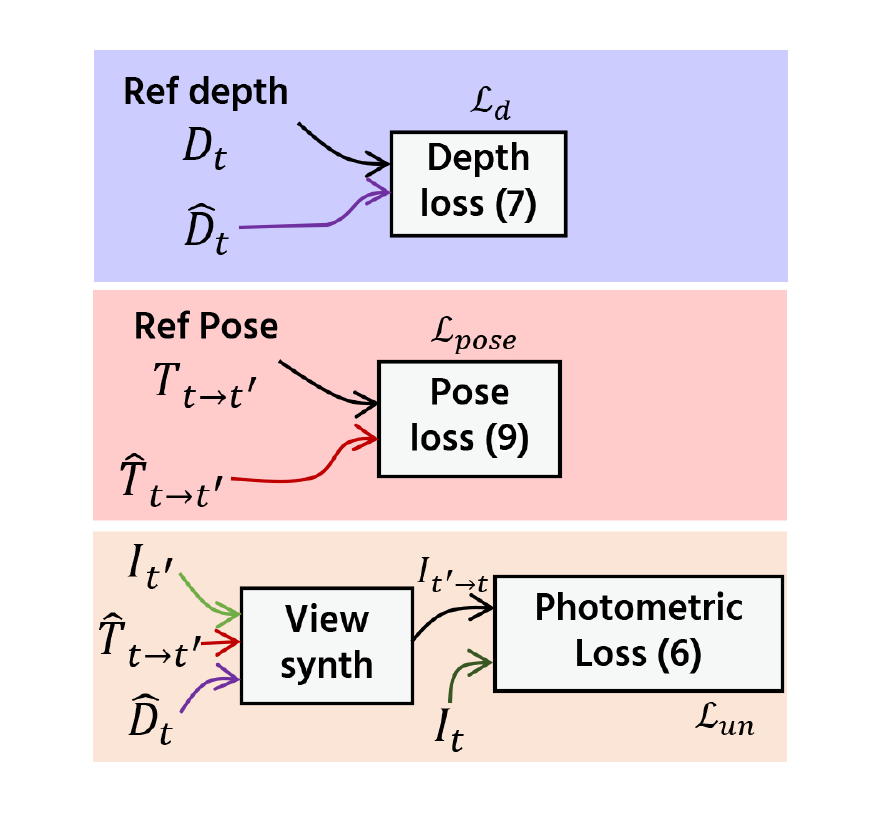}
    \vspace{-1mm}
    \caption{Direct depth and pose supervision~\cite{turan2018deep,widya2021self} + photometric loss.}
    \end{subfigure}
    \begin{subfigure}{0.24\textwidth}
    \centering
    \vspace{2mm}
    \includegraphics[trim=0 0 0 0,clip,width=\columnwidth]{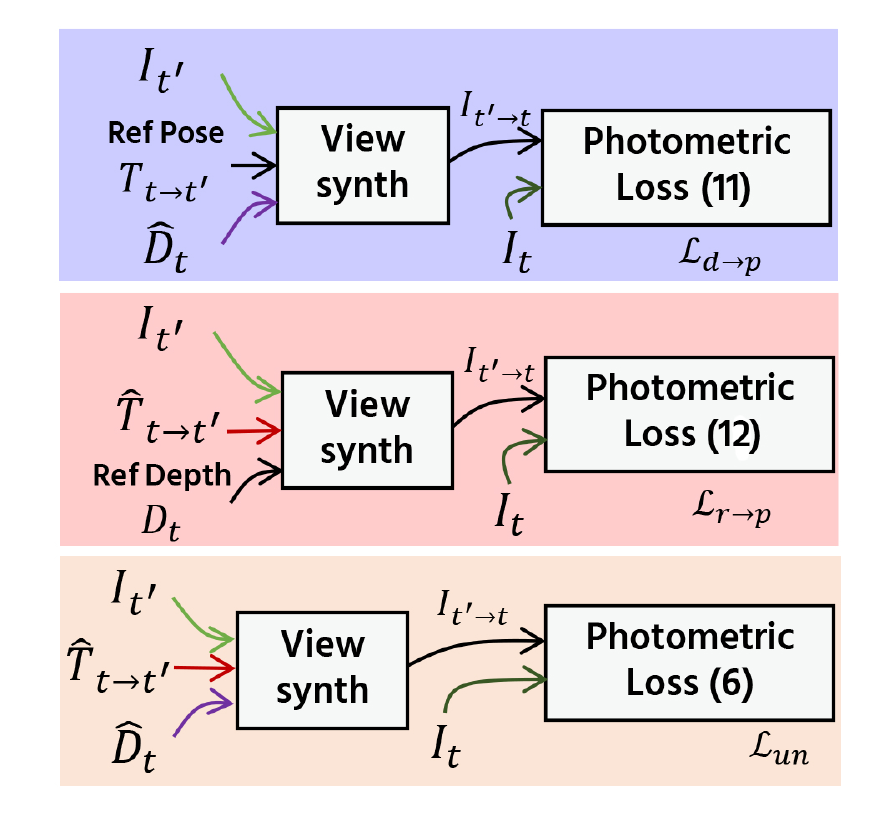}
    \vspace{-1mm}
    \caption{Proposed generalized photometric loss.}
    \end{subfigure}
    \caption{The network structure which consists of depth and pose estimation networks is shown in (a). Figures (b)-(d) show the comparison between the
    existing self-supervised photometric loss, the existing direct depth and pose supervision loss, and our proposed generalized photometric loss. In both~(c) and~(d), the loss in the purple-colored box is used for training the depth estimation network and the loss in the pink-colored box is used for training the pose estimation network. The existing depth and pose supervision approach trains the depth and the pose estimation networks by directly taking the Euclidean distance between the predicted depth and its reference and also between the predicted pose and its reference, respectively. This direct supervision approach needs balancing the weights for each loss term, which are difficult to search, because their physical meanings are different. In our proposed generalize loss, we adjusted our loss terms so that each of them has the same physical meaning, i.e., the photometric error. This generalization eliminates the need for the balancing weight search.}
    \label{fig:overall}
    \vspace{-3mm}
\end{figure*}

In this paper, we propose a supervised approach to simultaneously train depth and pose networks using consecutive images for monocular endoscopy of the stomach. To avoid the generalization problem between CG and real data, we apply the whole stomach 3D reconstruction pipeline~\cite{widya2019whole} to generate reference depth and pose from real endoscope data for network training. Additionally, we propose a novel loss generalization by unifying the depth and the pose losses into a photometric error loss for our supervised training to avoid the necessity of delicate weight balancing between the depth and the pose losses. Finally, we show that our supervised training with a novel generalized loss function has better performance than the existing direct depth and pose supervision.
Our method achieves up to 60fps at test time for depth and pose predictions.

\section{Materials and Methods}
Figure~\ref{fig:overall}(a) overviews the network structures used in our experiment and Figure 1(b)-(d) show how we train them using three different methods, i.e., an existing self-supervised photometric loss~\cite{godard2019digging}, an existing direct depth and pose losses~\cite{turan2018deep,widya2021self} combined with the photometric loss, and our proposed generalized photometric loss.
In this section, we firstly explain the endoscope dataset~(Section~\ref{sec:dataset}). We then review the existing self-supervision and direct supervision methods~(Section~\ref{sec:selfsupervised},~\ref{sec:direct}). Finally, we explain our proposed loss generalization~(Section~\ref{sec:proposed}).

\subsection{Training data generation} \label{sec:dataset}
In this work, we used the same endoscope video dataset from our previous work~\cite{widya2019whole}. We used six subjects' endoscope videos undergone general gastroendoscopy procedure.
We then extracted all the image frames from all the videos and used them as training and testing data. The experimental protocol was approved by the research ethics committee of Tokyo Institute of Technology and Nihon University Hospital. 

\begin{figure}[t!]
\centering
\begin{adjustbox}{max width = \columnwidth}
\begin{subfigure}{0.3\columnwidth}
    \centering
    \includegraphics[width=0.95\columnwidth]{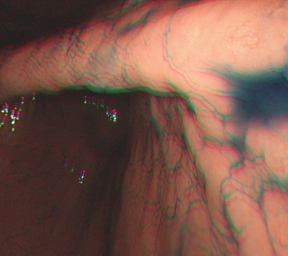} \\
\end{subfigure}
\begin{subfigure}{0.3\columnwidth}
    \centering
    \includegraphics[width=0.95\columnwidth]{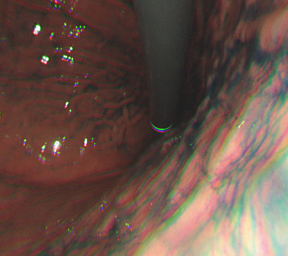} \\
\end{subfigure}
\begin{subfigure}{0.3\columnwidth}
    \centering
    \includegraphics[width=0.95\columnwidth]{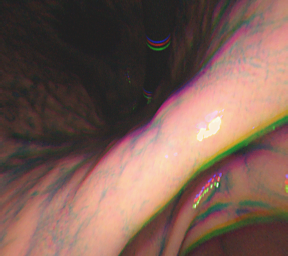} \\
\end{subfigure}
\end{adjustbox}

\begin{adjustbox}{max width = \columnwidth}
\begin{subfigure}{0.3\columnwidth}
    \centering
    \includegraphics[width=0.95\columnwidth]{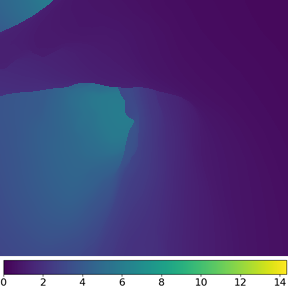} \\ 
\end{subfigure}
\begin{subfigure}{0.3\columnwidth}
    \centering
    \includegraphics[width=0.95\columnwidth]{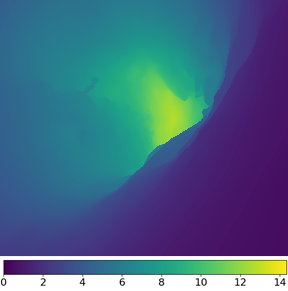} \\ 
\end{subfigure}
\begin{subfigure}{0.3\columnwidth}
    \centering
    \includegraphics[width=0.95\columnwidth]{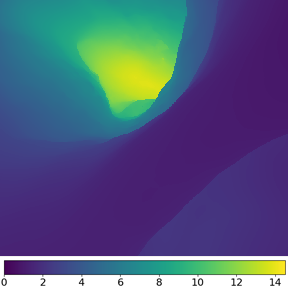} \\ 
\end{subfigure}
\end{adjustbox}

\caption{Some examples of the generated reference depth based on the estimated camera poses and the obtained whole stomach 3D model using~\cite{widya2019whole}. We can see that the generated reference depth images reflect the structures seen in the color images.}
\label{fig:gt_example}
\end{figure}

For the direct supervision~(Section~\ref{sec:direct}) and the proposed loss generalization~(Section~\ref{sec:proposed}) methods, we used the previously extracted real endoscope image frames to generate the reference depth and pose. For this purpose, we firstly applied the whole stomach 3D reconstruction pipeline~\cite{widya2019whole} and then extracted the reference depth and pose from the generated whole stomach 3D model and the estimated endoscope poses.
Figure~\ref{fig:gt_example} shows some of the RGB images and the generated reference depths. We used the generated reference depth and pose for both training and testing. 

\subsection{Self-supervised depth and pose estimation} \label{sec:selfsupervised}
Our depth and pose estimation networks are inspired by monodepth2 architecture~\cite{godard2019digging} shown in Figure~\ref{fig:overall}(a). It consists of two separate networks, each for depth and pose estimation purpose. Both networks are trained together to learn a view-synthesis problem, i.e., to predict the appearance of a target image given a view point of another image by minimizing its photometric error.

Let $I_t$ be a target frame and $I_{t'}$ be a source frame. The objective of the network is to minimize a photometric error \textit{pe} between a target frame and a warped source frame. 
In general, a photometric error \textit{pe} between two images $A$ and $B$ can be defined using pixel value difference (L1) and structural similarity index measure (SSIM)~\cite{wang2004image} such that
\begin{equation}
    pe(A, B) = \frac{\alpha}{2}(1 - \textrm{SSIM}(A, B))+(1-\alpha)\|A - B\|_1 
\end{equation}
where $\alpha$ is the balancing term between the L1 and the SSIM terms.
Let $\hat{D}_t$ be the predicted depth of the target frame $I_t$, $\hat{T}_{t \rightarrow t'}$ be the predicted relative pose from the target frame $I_t$ to the source frame $I_{t'}$, and $K$ be a calibrated camera intrinsic parameters. We then define the view synthesis (image warping) problem as
\begin{equation}
    I_{t' \rightarrow t} = \epsilon(I_{t'}, \pi(\hat{D}_t, \hat{T}_{t\rightarrow t'}, K))
\end{equation}
where $\pi$ is a function to project the pixel coordinate of target image $I_t$ in source image $I_{t'}$ and $\epsilon$ is a pixel sampling function based on the projected pixel coordinate given by $\pi$.
In our implementation, we used two consecutive frames as our source frames, \textit{i.e.}, $I_{t+1}$ and $I_{t-1}$. Instead of averaging the photometric error \textit{pe} for each warped source frame, we simply take the minimum such that the final pixel-wise photometric error can be expressed as
\begin{equation}
    \mathcal{L}_p = \min_{t'\in (t+1, t-1)} pe(I_t, I_{t' \rightarrow t})
    \label{eq:view_synth}
\end{equation}

To ensure that only the reliable pixels are optimized, we masked out the non-reliable pixel using the automask~\cite{godard2019digging} defined by a logical operation as
\begin{equation}
    \label{eq:static_mask}
    \mu = [\min_{t'\in (t+1, t-1)} pe(I_t, I_{t' \rightarrow t}) < \min_{t'\in (t+1, t-1)} pe(I_t, I_{t'})]
\end{equation}
We also used edge-aware smoothness so that there is no discontinuities in the predicted depth~\cite{godard2017unsupervised} that can be expressed as
\begin{equation}
    \mathcal{L}_s = |\partial_x \hat{d}^*_t|e^{-|\partial_x I_t|}+|\partial_y \hat{d}^*_t|e^{-|\partial_y I_t|}
\end{equation}
where $\partial_x$ and $\partial_y$ is the partial derivative on each $x$ and $y$ direction and $\hat{d}^*_t$ is the mean-normalized predicted inverse depth~\cite{wang2018learning}.
The final self-supervised loss function consists of the masked photometric error and the smoothness term as
\begin{equation}
    \label{eq:monodepthloss}
    \mathcal{L}_{un} = \frac{1}{N} \sum_{i}^{N} \mu^i \mathcal{L}_p^i + \lambda \mathcal{L}_s^i
\end{equation}
where $i$ represents a pixel index,  $N$ is the total number of the pixels, and $\lambda$ is the balancing weight between the photometric error and the depth smoothness loss.  

\subsection{Supervised depth and pose estimation} \label{sec:direct}

To supervise the depth estimation network, we followed the method~\cite{alhashim2018high} and used inverse depth $d$ instead of depth $D$. We followed a standard inverse depth error loss function which compares the inverse depth prediction $\hat{d}_t$ and its reference inverse depth $d_t$. It consists of three sub-components which can be formulated as follows
\begin{subequations}
\begin{align}
    \mathcal{L}_{d} &= |\hat{d}_{t} - d_{t}| \label{eq:L1_loss}, \\
    \mathcal{L}_{g} &= |\partial_x(\hat{d}_{t}, d_{t})| \label{eq:grad}+|\partial_y(\hat{d}_{t}, d_{t})|, \\
    \mathcal{L}_{\textrm{SSIM}} &= \frac{1 - \textrm{SSIM}(\hat{d}_t,d)}{2} \label{eq:SSIM}
\end{align}
\end{subequations}
The total loss for depth supervision can finally be expressed as
\begin{align}
    \mathcal{L}_{d_t} = \frac{1}{N} \sum_{i}^{N} 0.1 \mathcal{L}_{d}^i + \mathcal{L}_{g}^i + \mathcal{L}_{\textrm{SSIM}}^i.
\end{align}

For the pose estimation supervision, the common practice is to directly supervise the pose estimation network by measuring Euclidean distance between the predicted relative pose and its reference pose~\cite{kendall2015posenet, ummenhofer2017demon,turan2018deep}, i.e.,

\begin{equation}
    \label{eq:pose}
    \mathcal{L}_{pose}^{t\rightarrow t'} = \zeta\|\hat{\textbf{x}}_{t\rightarrow t'} - \textbf{x}_{t\rightarrow t'}\|_2 + \theta \|\hat{\textbf{r}}_{t\rightarrow t'} - \textbf{r}_{t\rightarrow t'}\|_2 
\end{equation}
where $\textbf{x}$ is the translation vector component and $\textbf{r}$ is the rotation vector components in the axis-angle representation from the relative pose $T_{t\rightarrow t'}$. The translation and rotation terms are balanced by $\zeta$ and $\theta$ as weights.
To tie $\mathcal{L}_d$ and $\mathcal{L}_{pose}$ together, a photometric loss $\mathcal{L}_{p}$ is added to the direct supervision loss. Finally, the total supervised loss can be expressed as
\begin{equation}
    \label{eq:supervised}
    \mathcal{L}_{su} = \frac{1}{N} \sum_i^N (\psi \mu^i \mathcal{L}_p^i + \gamma \mathcal{L}_{d_t}^i) + \sum_{j\in t'} \mathcal{L}_{pose}^{t\rightarrow j}
\end{equation}
where $\psi$ and $\gamma$ are balancing weights for depth and pose losses.

\subsection{Loss generalization} \label{sec:proposed}

Since each of the components in~(\ref{eq:supervised}) in the commonly used supervised loss has different physical meaning, the weight of each component has to be carefully selected, which is very difficult and usually performed in an empirical manner. It is also common that different weight balancing is needed for different kinds of environment such as outdoor and indoor scenes~\cite{kendall2015posenet}. To address this limitation, we propose a novel depth and pose supervision loss function by generalizing the depth and the pose errors into the same physical meaning, i.e, a photometric error.

\begin{figure*}
\begin{center}
    \begin{tikzpicture}[xscale=2.8,yscale=2.5]
      \node at (0,3.62) {\footnotesize Input image};
      \node at (1,3.62) {\shortstack{\footnotesize Reference\\ \footnotesize depth}};
      \node at (2,3.62) {\footnotesize Self-supervised~\cite{godard2019digging}};
      \node at (3,3.62) {\shortstack{\footnotesize Direct\\ \footnotesize supervision~\cite{turan2018deep,widya2021self}}};
      \node at (4,3.62) {\shortstack{\footnotesize Generalized\\ \footnotesize loss (Ours)}};
      
      \node at (0,3) {\includegraphics[width=0.14\textwidth]{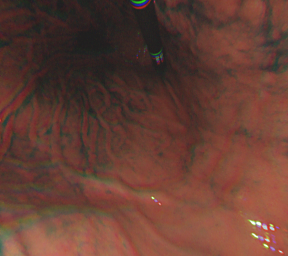}};
      \node at (0,2) {\includegraphics[width=0.14\textwidth]{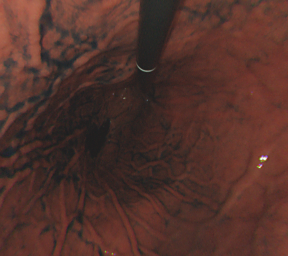}};
      \node at (0,1) {\includegraphics[width=0.14\textwidth]{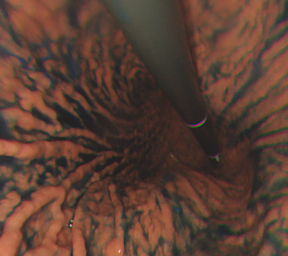}};
      \node at (0,0) {\includegraphics[width=0.14\textwidth]{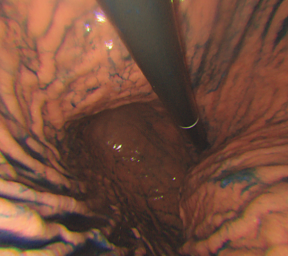}};
      
      \node at (1,3) {\includegraphics[width=0.14\textwidth]{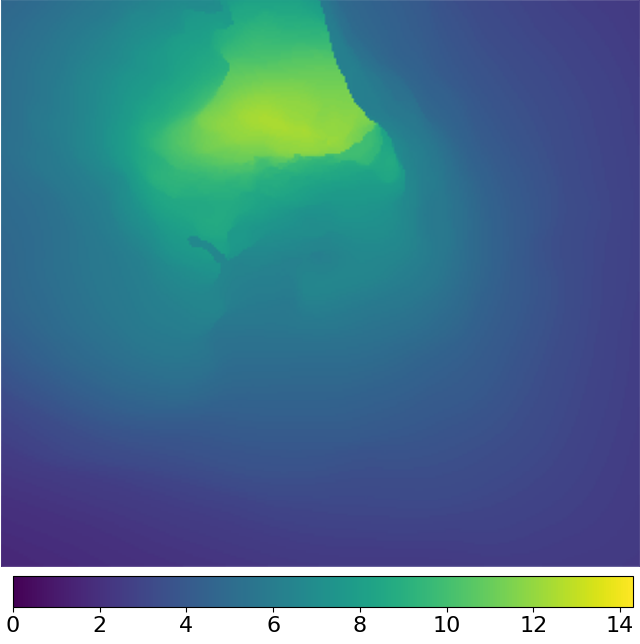}};
      \node at (1,2) {\includegraphics[width=0.14\textwidth]{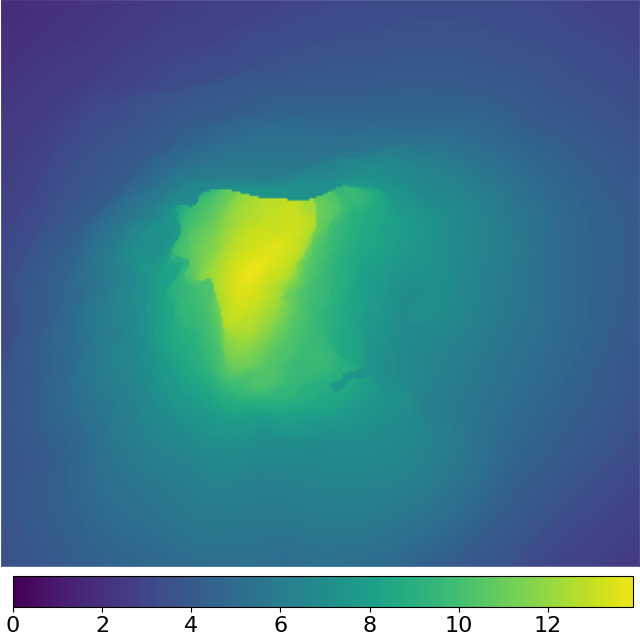}};
      \node at (1,1) {\includegraphics[width=0.14\textwidth]{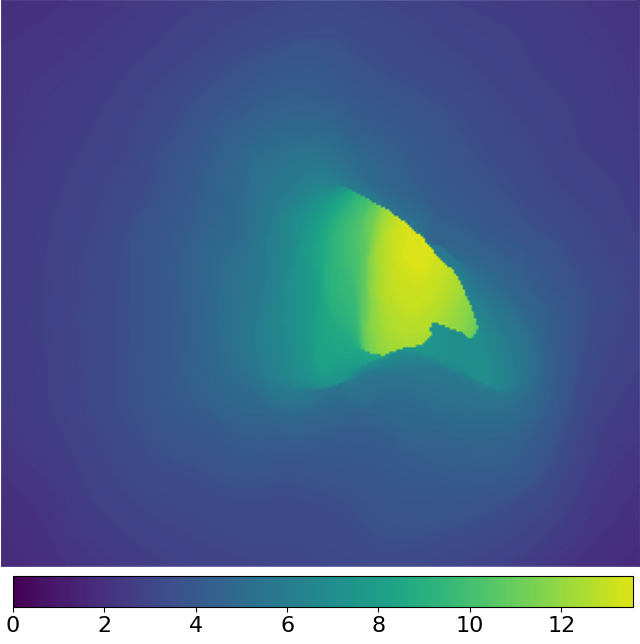}};
      \node at (1,0) {\includegraphics[width=0.14\textwidth]{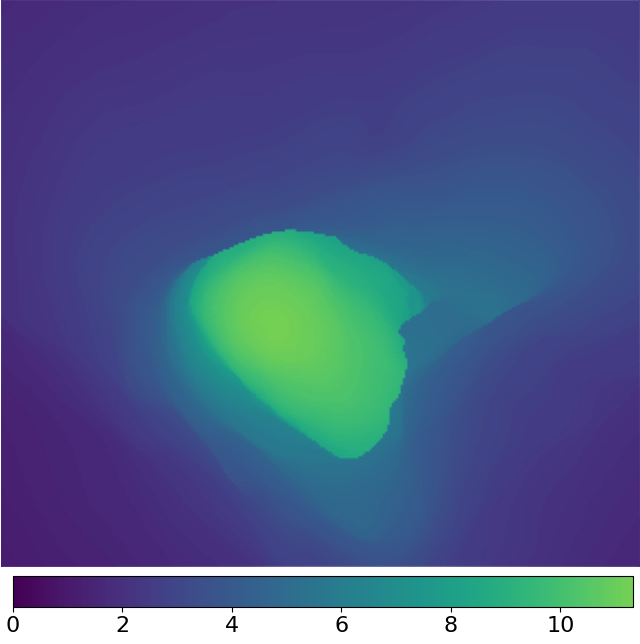}};
      
      \node at (2,3) {\includegraphics[width=0.14\textwidth]{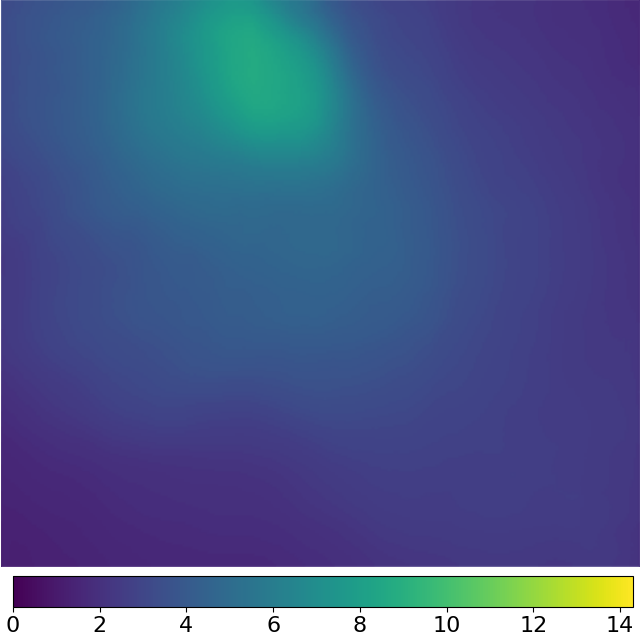}};
      \node at (2,2) {\includegraphics[width=0.14\textwidth]{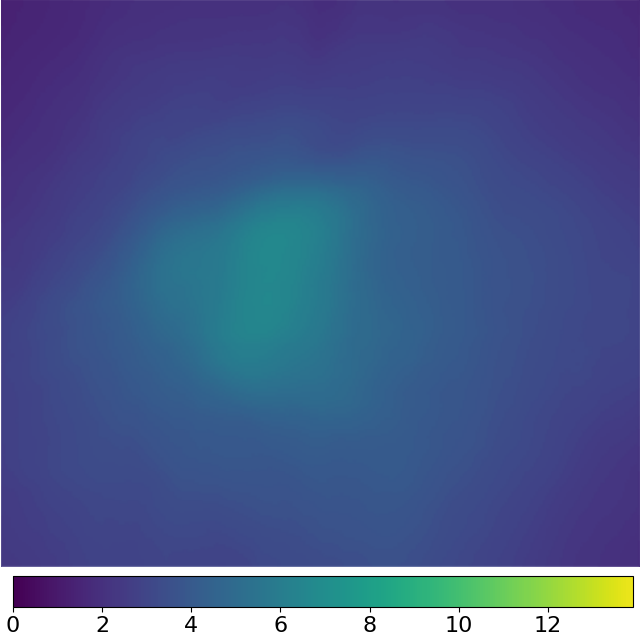}};
      \node at (2,1) {\includegraphics[width=0.14\textwidth]{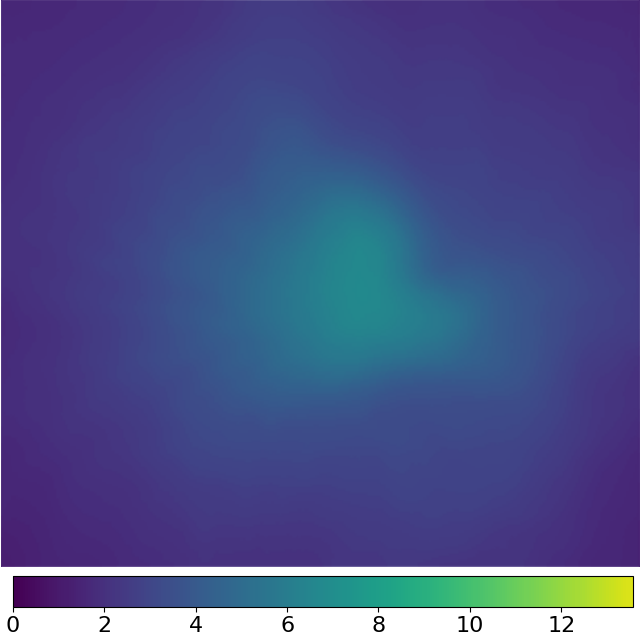}};
      \node at (2,0) {\includegraphics[width=0.14\textwidth]{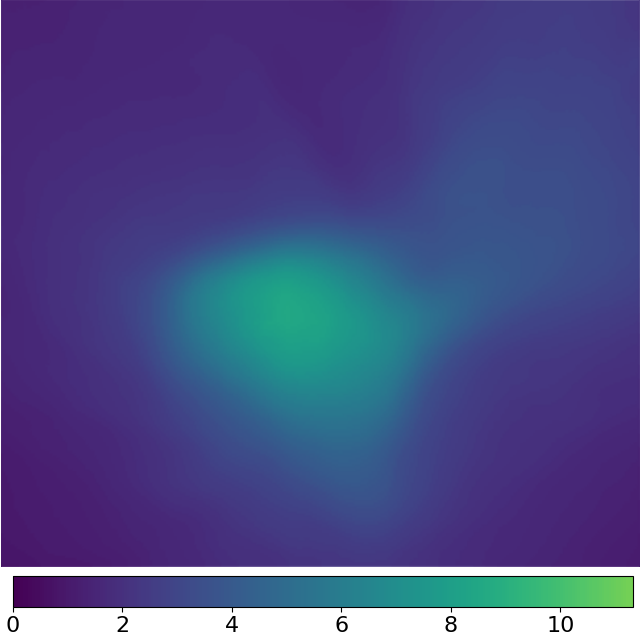}};
      
      \node at (3,3) {\includegraphics[width=0.14\textwidth]{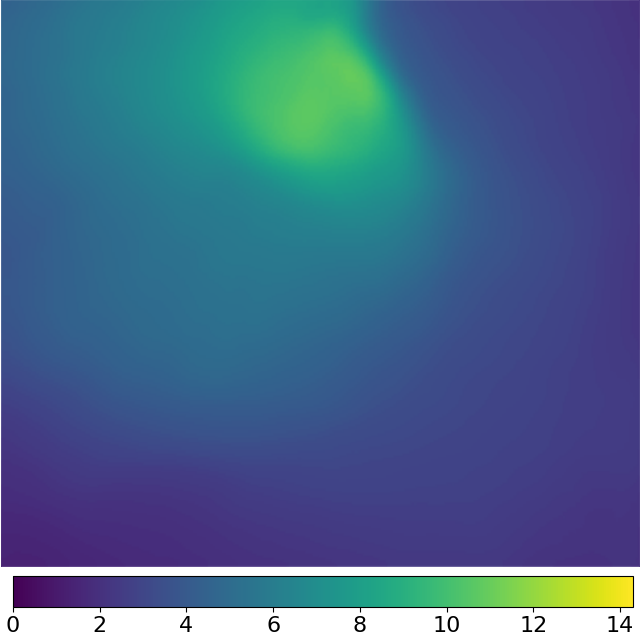}};
      \node at (3,2) {\includegraphics[width=0.14\textwidth]{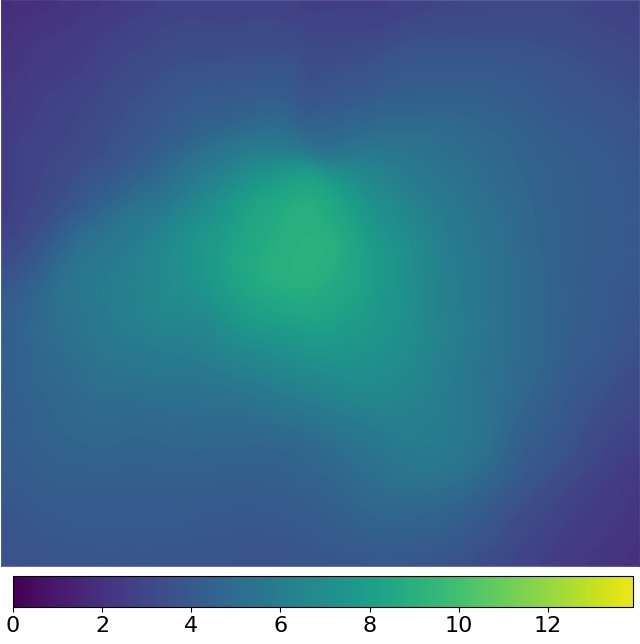}};
      \node at (3,1) {\includegraphics[width=0.14\textwidth]{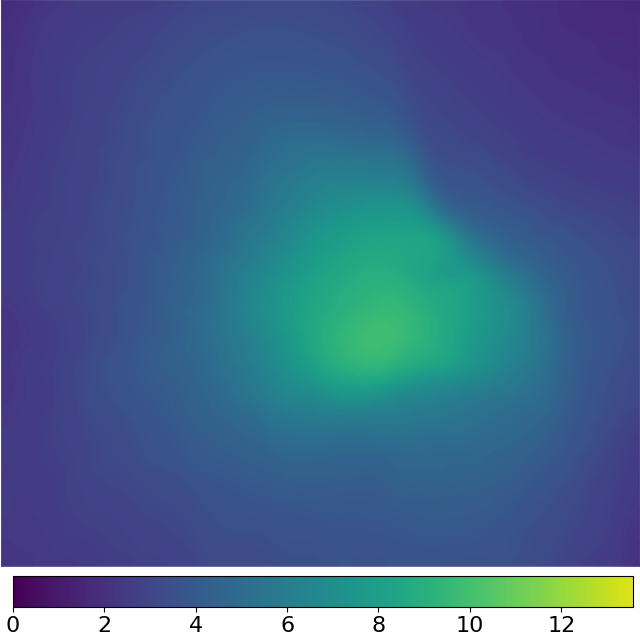}};
      \node at (3,0) {\includegraphics[width=0.14\textwidth]{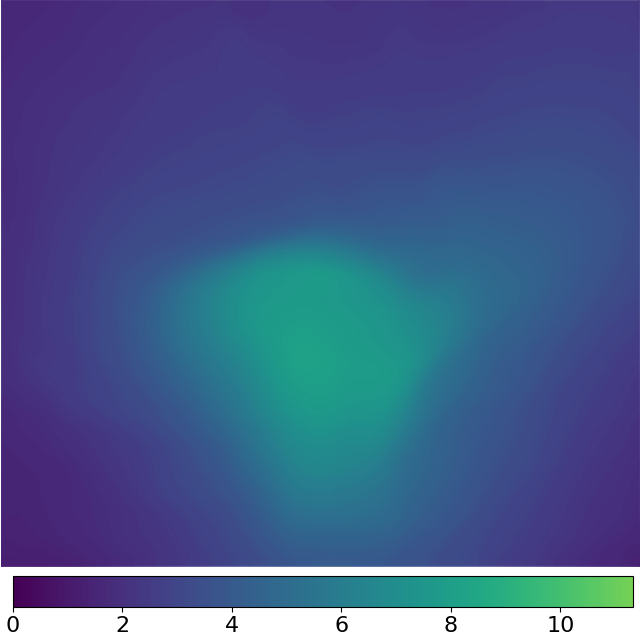}};
      
      \node at (4,3) {\includegraphics[width=0.14\textwidth]{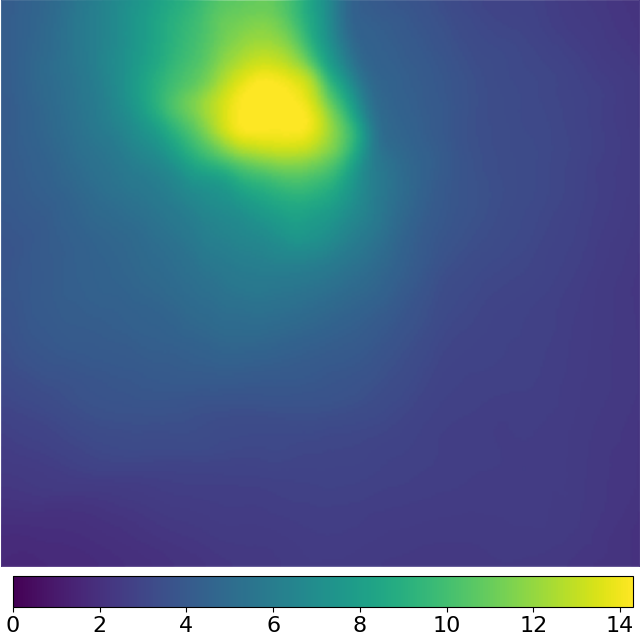}};
      \node at (4,2) {\includegraphics[width=0.14\textwidth]{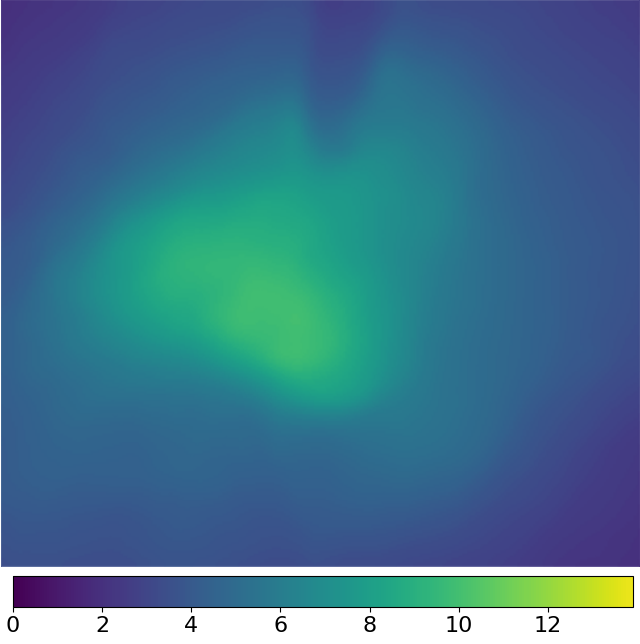}};
      \node at (4,1) {\includegraphics[width=0.14\textwidth]{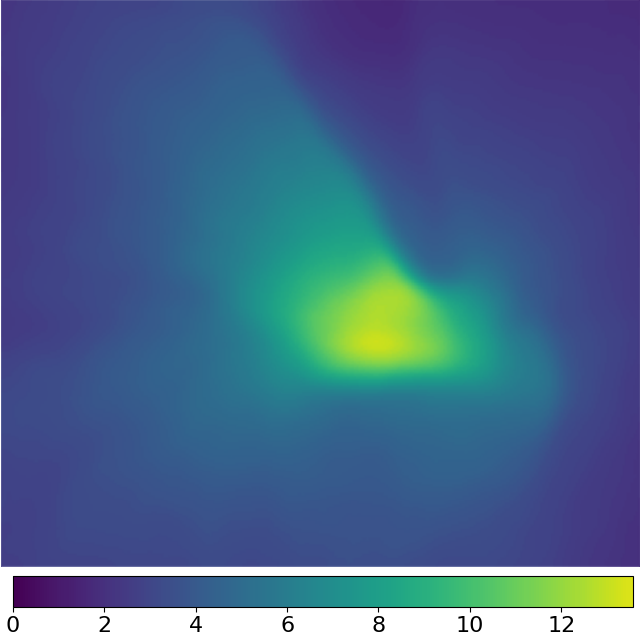}};
      \node at (4,0) {\includegraphics[width=0.14\textwidth]{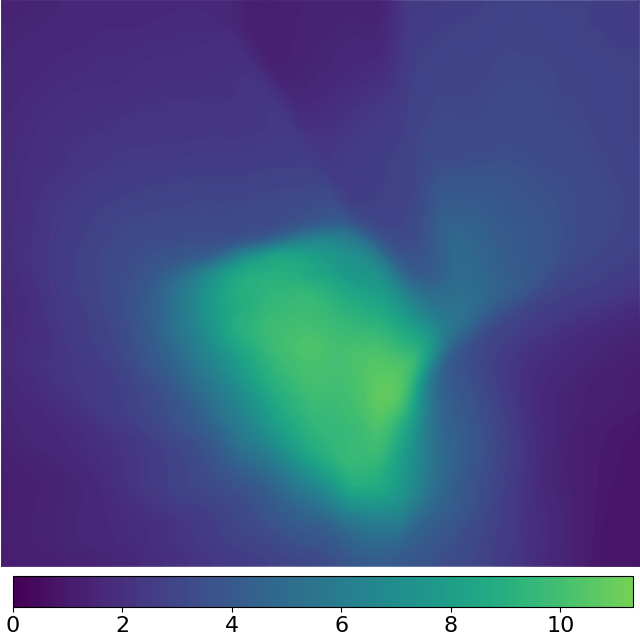}};
      
    \end{tikzpicture}
\end{center}
\vspace{-4mm}
\caption{Some examples of depth estimation results. Here we show the RGB images for better visualization, though we actually used red channel images as the input of the network according to the finding in~\cite{widya2019whole}. We compare the depth prediction results of the self-supervision~\cite{godard2019digging}, the direct supervision~\cite{turan2018deep,kendall2015posenet}, and our proposed generalized loss supervision. As we can see, our proposed method not only estimates closer depth to the reference, but also better estimates the structures and the boundaries, including the endoscope rod. In some cases, the direct supervision results are too smooth.}
\label{fig:depthresult}
\vspace{-4mm}
\end{figure*}
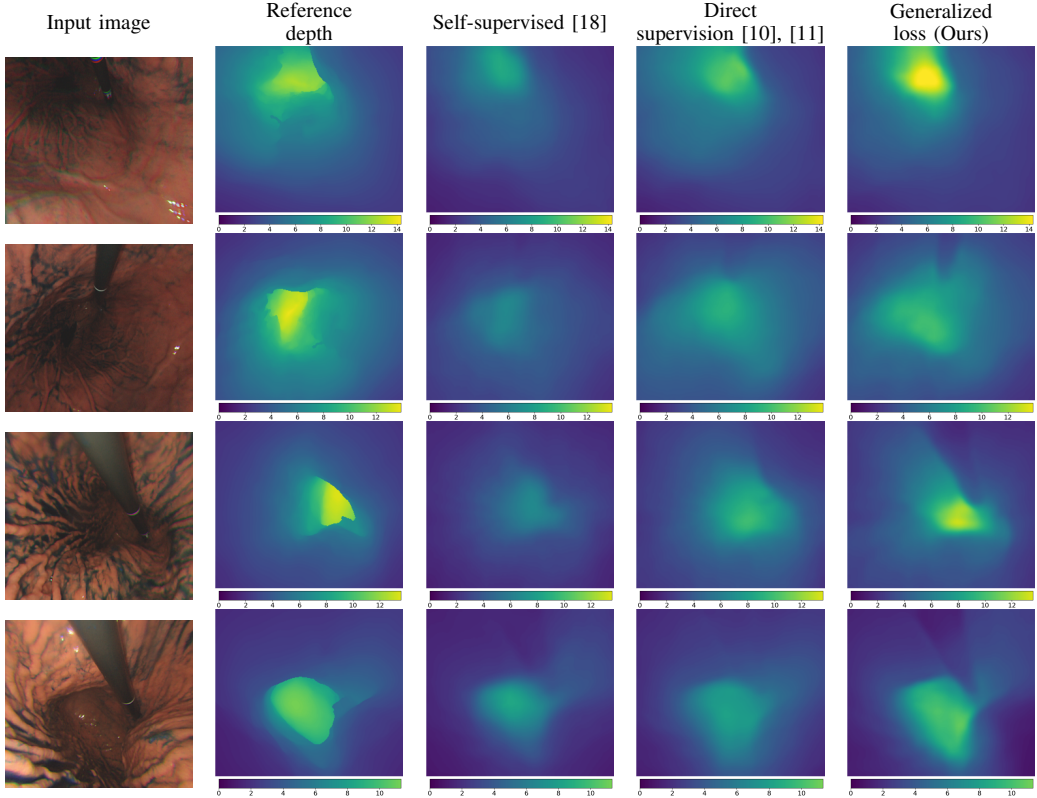

As illustrated in Figure 1(d), in order to generalize the loss into a photometric error, we use the reference relative pose $T_{t \rightarrow t'}$ to train the depth estimation network by optimizing the predicted depth $\hat{D}_t$ such that
\begin{equation}
    \label{eq:ldp}
    \mathcal{L}_{d\rightarrow p} = \min_{t'\in (t+1, t-1)} pe(I_t, \epsilon(I_{t'}, \pi(\hat{D}_t, T_{t\rightarrow t'}, K)))
\end{equation}
Conversely, we use the reference depth $D_t$ to train the pose estimation network by optimizing the predicted relative pose $\hat{T}_{t \rightarrow t'}$ such that
\begin{equation}
    \label{eq:lrp}
    \mathcal{L}_{r\rightarrow p} = \min_{t'\in (t+1, t-1)} pe(I_t, \epsilon(I_{t'}, \pi(D_t, \hat{T}_{t\rightarrow t'}, K)))
\end{equation}
The term described in~(\ref{eq:ldp}) can be defined as \textit{depth loss w.r.t reference pose as photometric loss} while the term described in~(\ref{eq:lrp}) as \textit{pose loss w.r.t reference depth as photometric loss}. We also calculate the reliable pixel masks, $\mu_{d\rightarrow p}$ and $\mu_{r\rightarrow p}$, for each $\mathcal{L}_{d\rightarrow p}$ and $\mathcal{L}_{r\rightarrow p}$ respectively using the same principle as~(\ref{eq:static_mask}).

Combining~(\ref{eq:ldp}) and~(\ref{eq:lrp}) with~(\ref{eq:view_synth}) to tie the depth and the pose network training together, we can write the final loss function as
\begin{equation}
\begin{split}
    \mathcal{L}_{gen} &= \frac{1}{N} \sum_i^N (\mu_{d\rightarrow p}^i \mathcal{L}_{d\rightarrow p}^i + \mu_{r\rightarrow p}^i \mathcal{L}_{r\rightarrow p}^i \\ & + \underbrace{\mu^i \mathcal{L}_{p}^i + \lambda \mathcal{L}_s^i}_{\mathcal{L}_{un}~(\ref{eq:monodepthloss})})
\end{split}
\end{equation}
which eliminates the intricate search of balancing weights for the depth and the pose loss terms.

\section{Results and Discussion}
\subsection{Implementation details}
Following~\cite{godard2019digging}, we used ResNet-18 architecture~\cite{he2016deep} for our depth and pose estimation networks. We simultaneously trained our depth and pose estimation networks using a single NVIDIA GeForce GTX 1080Ti GPU. Our networks were trained for 100 epochs with the learning rate of $10^{-4}$ with the decay factor of $10^{-1}$ after 50 epochs. We set the term weights for the self-supervised and the generalized loss training as $\alpha=0.85$ and $\lambda = 0.001$. 
Additionally, we set the extra balancing weights for the direct supervision as $\gamma = 30$, $\zeta = \psi = 15$, and $\theta = 160$.

We divided six subjects into four subjects for training (Subjects 3-6, 9000 images) and two subject for testing (Subjects 1-2, 2350 images). The image resolution is $288 \times 256$ pixels. Following the finding of our previous research to tackle the color channel misalignment problem~\cite{widya2019whole}, we used only red channel images to train the networks.

\begin{figure}
\begin{center}
    \begin{tikzpicture}[xscale=4,yscale=3.2]
      \node at (0,1.2) {\includegraphics[width=0.32\columnwidth]{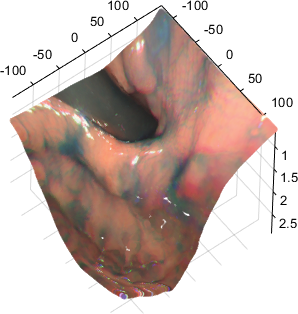}};
      \node at (1,1.2) {\includegraphics[width=0.32\columnwidth]{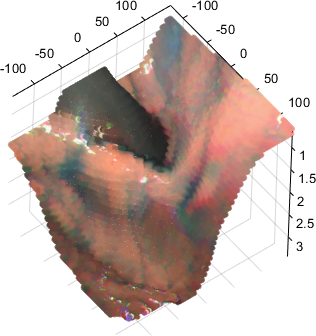}};
      \node at (-.5,1.5) {\includegraphics[width=0.15\columnwidth]{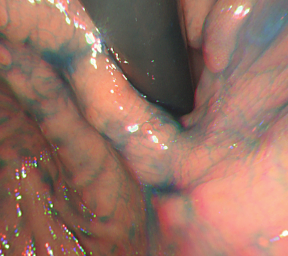}};
      \node at (-.5,1.72) {\footnotesize Input image};
      \node at (-.5,1.0) {\includegraphics[width=0.15\columnwidth]{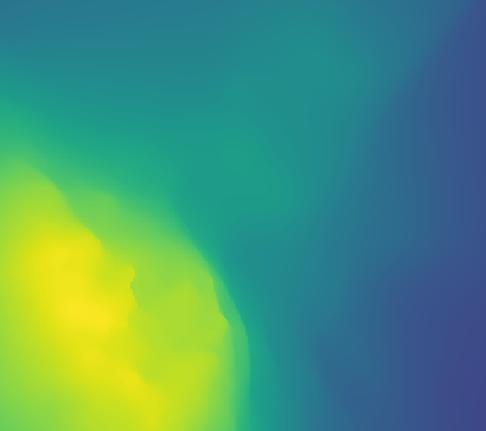}};
      \node at (-.5,1.22) {\footnotesize Ref. depth};
      \node at (0.55,1.5) {\includegraphics[width=0.15\columnwidth]{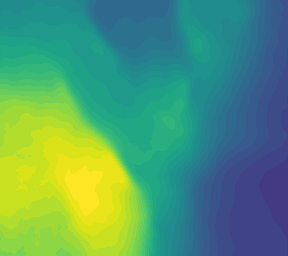}};
      \node at (0.55,1.72) {\footnotesize Pred. depth};
      
      \node at (0,0.2) {\includegraphics[width=0.32\columnwidth]{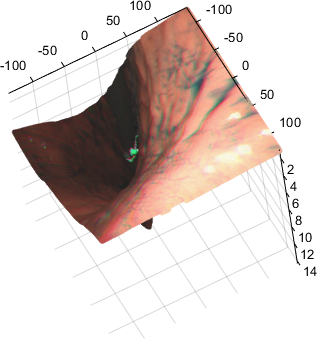}};
      \node at (1,0.2) {\includegraphics[width=0.32\columnwidth]{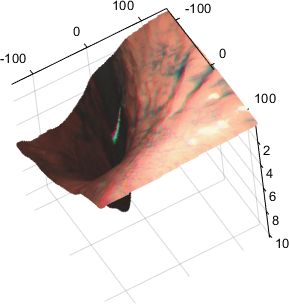}};
      \node at (-.5,0.5) {\includegraphics[width=0.15\columnwidth]{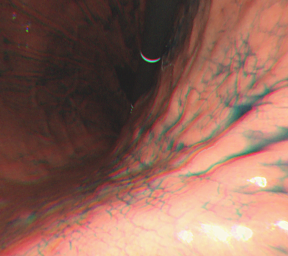}};
      \node at (-.5,0.72) {\footnotesize Input image};
      \node at (-.5,0.01) {\includegraphics[width=0.15\columnwidth]{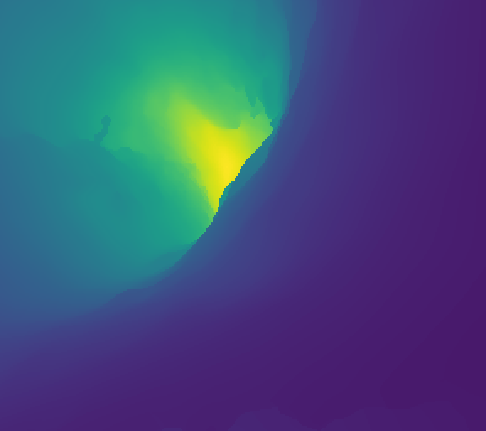}};
      \node at (-.5,0.22) {\footnotesize Ref. depth};

      \node at (.55,0.4) {\includegraphics[width=0.15\columnwidth]{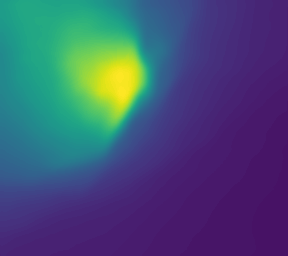}};
      \node at (.55,.62) {\footnotesize Pred. depth};
      
      \node at (-.1,-.25) {(a) Reference depth};
      \node at (0.9,-.25) {(b) Generalized loss (Ours)};
    \end{tikzpicture}
\end{center}
\vspace{-2mm}
\caption{Comparison of the generated 3D point clouds using the reference depth and the predicted depth by our proposed method. As we can see, both the depth and the structure from the predicted depth are close to the ones generated from the reference depth.}
\label{fig:pcl}
\end{figure}
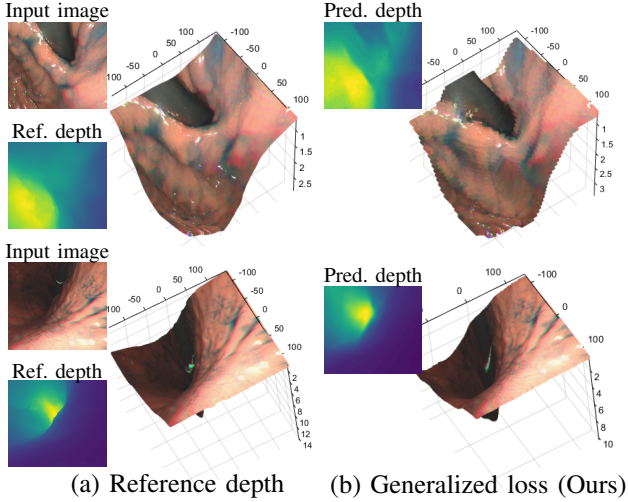

\subsection{Depth estimation evaluation}
Figure~\ref{fig:depthresult} shows the subjective evaluation results. We evaluated the relative depth error and the depth accuracy as
\begin{itemize}
    \item Relative error: $\frac{|\hat{D}_{t} - D_{t}|}{D_{t}}$
    \item Depth accuracy: $\delta = \max(\frac{D_{t}}{\hat{D}_{t}}, \frac{\hat{D}_{t}}{D_{t}})$
\end{itemize}
where $D_t$ is the reference depth and $\hat{D}_t$ is the predicted depth. For the relative error, we calculated the errors for all the pixels of all the frames and evaluated the mean, the maximum, and the median values. For the depth accuracy evaluation, we measured the ratio between the number of pixels that have a lower error than a threshold controlled by $k$ (i.e., {$\delta < 1.25^k$}) and the total number of the pixels.

\renewcommand{\arraystretch}{1.4}
\begin{table}
\centering
\vspace{-2mm}
\caption{Depth estimation objective evaluation.}
\label{tab:depth_eval}
\resizebox{\columnwidth}{!}{%
\begin{tabular}{ccccccc}
\hline
\multicolumn{1}{c|}{}                                                                      & \multicolumn{3}{c|}{Accuracy $\uparrow$}                                         & \multicolumn{3}{c}{Relative errors $\downarrow$}                       \\ \hline
\multicolumn{1}{c|}{Method}                                                                & \begin{tabular}[c]{@{}c@{}}$\delta$\\$<1.25^1$\end{tabular} & \begin{tabular}[c]{@{}c@{}}$\delta$\\$<1.25^2$\end{tabular} & \multicolumn{1}{c|}{\begin{tabular}[c]{@{}c@{}}$\delta$\\$<1.25^3$\end{tabular}} & mean & max & median                           \\ \hline
\multicolumn{7}{c}{Test on Subject 1}                                                                                                                                                                                         \\ \hline
\multicolumn{1}{c|}{Self-supervised~\cite{godard2019digging}}                                                            & 0.374          & 0.703          & \multicolumn{1}{c|}{0.838}          & 0.635 &8.812 & 0.286       \\
\multicolumn{1}{c|}{Direct supervision~\cite{turan2018deep,widya2021self}} & 0.525          & \textbf{0.814}          & \multicolumn{1}{c|}{0.900}          & 0.432 &6.628 & \textbf{0.209}          \\
\multicolumn{1}{c|}{Generalized loss (Ours)}   & \textbf{0.540} & 0.804 & \multicolumn{1}{c|}{\textbf{0.902}} & \textbf{0.416} & \textbf{5.445} & 0.212 \\ \hline
\multicolumn{7}{c}{Test on Subject 2}                                                                                                                                                                                         \\ \hline
\multicolumn{1}{c|}{Self-supervised~\cite{godard2019digging}}                                                            & 0.477          & 0.767          & \multicolumn{1}{c|}{0.867}  & 0.472 & 8.673 & 0.227          \\
\multicolumn{1}{c|}{Direct supervision~\cite{turan2018deep,widya2021self}} & 0.536          & 0.806          & \multicolumn{1}{c|}{0.910}  & 0.349 & 6.717 & \textbf{0.210}          \\
\multicolumn{1}{c|}{Generalized loss (Ours)}   & \textbf{0.579} & \textbf{0.822} & \multicolumn{1}{c|}{\textbf{0.916}} & \textbf{0.336} & \textbf{6.632} & 0.213 \\ \hline

\multicolumn{7}{c}{Test on Training data}                                                                       \\ \hline
\multicolumn{1}{c|}{Self-supervised~\cite{godard2019digging}}                                                            & \multicolumn{1}{c}{0.565}          & \multicolumn{1}{c}{0.819}          & \multicolumn{1}{c|}{0.912}          & \multicolumn{1}{c}{0.342} & \multicolumn{1}{c}{3.602} & \multicolumn{1}{c}{0.204}          \\
\multicolumn{1}{c|}{Direct supervision~\cite{turan2018deep,widya2021self}} & \multicolumn{1}{c}{\textbf{0.961}} & \multicolumn{1}{c}{\textbf{0.992}} & \multicolumn{1}{c|}{\textbf{0.996}} & \multicolumn{1}{c}{\textbf{0.064}} &  \multicolumn{1}{c}{\textbf{0.465}} &\multicolumn{1}{c}{\textbf{0.059}} \\
\multicolumn{1}{c|}{Generalized loss (Ours)}   & \multicolumn{1}{c}{0.791}          & \multicolumn{1}{c}{0.916}          & \multicolumn{1}{c|}{0.956}          & \multicolumn{1}{c}{0.168} & \multicolumn{1}{c}{1.313} & \multicolumn{1}{c}{0.114}          \\ \hline
\multicolumn{1}{l}{\scalebox{0.9}{*Numbers are dimensionless}}
\end{tabular}%
}
\vspace{-8mm}
\end{table}

Table~\ref{tab:depth_eval} shows the objective evaluation results. Since the predicted depth is only up to scale, we scaled the predicted depth by minimizing the average RMSE for the entire sequence.
From Table~\ref{tab:depth_eval}, we can see that our proposed method has better performance compared to the self-supervised method by a fair margin. In addition, our proposed method generally shows better performance compared to the direct supervision method. Even though our proposed method comes seconds in the median relative error, the values are very close. In addition to the testing on the test data (Subject 1 and 2), we also tested each of the trained networks on the training data. As we can see, the direct supervision has the best results for this evaluation. However, it can be noticed that the performance of the direct supervision on the test data falls sharply compared to its performance on the training data. It shows that the depth estimation network trained with the direct supervision has poor generalization to the data that have never been seen during the training.

Figure~\ref{fig:pcl} shows the resulting color point clouds by fusing a single image with its predicted depth. As we can see, the resulting point cloud from our predicted depth is very close to the point cloud from the reference depth.

\begin{table}
\centering
\caption{Pose estimation objective evaluation.}
\label{tab:pose_eval}
\resizebox{\columnwidth}{!}{%
\begin{tabular}{ccccccc}
\hline
\multicolumn{1}{c|}{}                                                                      & \multicolumn{3}{c|}{Rotation error $\downarrow$}                                   & \multicolumn{3}{c}{Translation error $\downarrow$}            \\ \hline
\multicolumn{1}{c|}{Method}                                                                & mean           & max            & \multicolumn{1}{c|}{median}            & mean           & max            & median            \\ \hline
\multicolumn{7}{c}{Test on Subject 1}                                                                                                                                                                                         \\ \hline
\multicolumn{1}{c|}{Self-supervised~\cite{godard2019digging}}                                                            & 0.562          & 0.809          & \multicolumn{1}{c|}{0.519}          & 0.253          & 0.531          & 0.211          \\
\multicolumn{1}{c|}{Direct supervision~\cite{turan2018deep, widya2021self}} & 0.579          & 0.887          & \multicolumn{1}{c|}{0.539}          & 0.274          & 0.555          & 0.226          \\
\multicolumn{1}{c|}{Generalized loss (Ours)}   & \textbf{0.458} & \textbf{0.714} & \multicolumn{1}{c|}{\textbf{0.426}} & \textbf{0.222} & \textbf{0.471} & \textbf{0.178} \\ \hline
\multicolumn{7}{c}{Test on Subject 2}                                                                                                                                                                                         \\ \hline
\multicolumn{1}{c|}{Self-supervised~\cite{godard2019digging}}                                                            & 0.581          & 0.802          & \multicolumn{1}{c|}{0.564} & 0.283          & 0.587          & 0.239          \\
\multicolumn{1}{c|}{Direct supervision~\cite{turan2018deep, widya2021self}} & 0.606          & 0.836          & \multicolumn{1}{c|}{0.588}          & 0.296          & 0.578          & 0.243          \\
\multicolumn{1}{c|}{Generalized loss (Ours)}   & \textbf{0.517} & \textbf{0.742} & \multicolumn{1}{c|}{\textbf{0.491}}          & \textbf{0.246} & \textbf{0.495} & \textbf{0.206} \\ \hline

\multicolumn{7}{c}{Test on Training data}                                                                              \\ \hline
\multicolumn{1}{c|}{Self-supervised~\cite{godard2019digging}}                                                            & 0.554          & 0.769          & \multicolumn{1}{c|}{0.511}          & 0.276          & 0.585          & 0.231          \\
\multicolumn{1}{c|}{Direct supervision~\cite{turan2018deep, widya2021self}} & \textbf{0.195} & \textbf{0.324} & \multicolumn{1}{c|}{\textbf{0.172}} & \textbf{0.116} & \textbf{0.265} & \textbf{0.093} \\
\multicolumn{1}{c|}{Generalized loss (Ours)}   & 0.385          & 0.544          & \multicolumn{1}{c|}{0.355}          & 0.182          & 0.371          & 0.154          \\ \hline
\multicolumn{1}{l}{\scalebox{0.9}{*Numbers are dimensionless}}
\end{tabular}%
}
\vspace{-6mm}
\end{table}

\subsection{Pose estimation evaluation}
For pose estimation evaluation, we first split the full sequences of Subject 1 and 2 into the groups of 150 consecutive frames. The predicted poses were then aligned with the reference poses using Umeyama transform~\cite{umeyama1991least}.
We then used absolute pose error (APE) to evaluate the translation and rotation components of the predicted poses $\hat{P} \in \textrm{SE(3)}$ against the reference poses $P \in \textrm{SE(3)}$. Given the absolute relative pose between a pair of predicted pose and its ground truth $E = P^{-1}\hat{P}$, the translation and the rotation errors can be defined as
\begin{equation}
    APE_{rot} = \|\textrm{rot}(E) - I_{3\times3}\|_F
\end{equation}
\begin{equation}
    APE_{trans} = \|\textrm{trans}(E)\|_2
\end{equation}
where $\|.\|_F$ is Frobenius norm. We then averaged all the obtained APEs over all of the evaluation points. The objective evaluation results can be seen in Table~\ref{tab:pose_eval}.

From Table~\ref{tab:pose_eval}, we can see that, based on the evaluation on the test data, our proposed generalized loss has better performance compared to the self-supervised and the direct supervision methods. Even though it is evident that the direct supervision has the best result when tested on the training data, its performance drops sharply when tested on the test data. This characteristic is consistent with the results previously shown in the depth estimation evaluation. This is because that the direct supervision losses induce poor generalization performance. In addition, even without the intricate search of the term-balancing weights, our method could achieve the best result for the test data. 

\begin{figure}
    \centering
    \includegraphics[width=0.75\columnwidth]{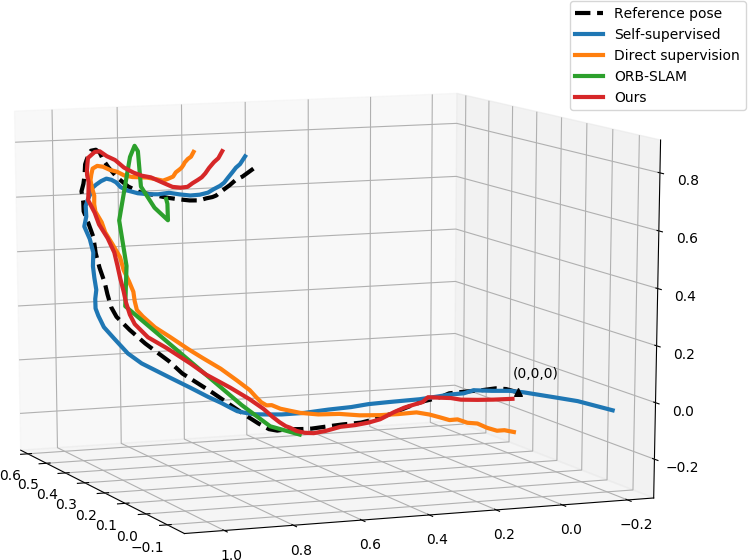}
    \caption{The predicted trajectory for a sample sequence. As we can see, our prediction result is the closest to the reference. ORB-SLAM~\cite{mur2015orb} could only predict the poses of 16 frames among 150 input frames.}
    \vspace{-6mm}
    \label{fig:trajectories}
\end{figure}

Finally, we show the trajectory prediction results in Figure~\ref{fig:trajectories}, including the trajectory prediction result from ORB-SLAM~\cite{mur2015orb}. As we can see, our prediction result is the closest to the reference, while ORB-SLAM could only predict the poses of 16 frames among the 150 input frames.

\section{Conclusions}
In this paper, we have presented a novel generalized photometric loss for learning-based depth and pose estimation with monocular endoscopy. Compared to commonly used direct depth and pose supervision losses, which have different physical meanings for each loss term, we have proposed the generalized loss so that each of the loss terms has the same physical meaning, which is a photometric error. We have experimentally shown that our generalized loss supervision performs better than the direct depth and pose supervision without the need for an intricate search of term-balancing weights. We have also found that the generalization performance from train to test data of our proposed method is better than that of the direct supervision. In future work, we plan to fuse both depth and pose predictions from multiple frames for real-time 3D reconstruction and use the light source information to improve the depth and pose estimation. Additional related results be accessed from (\textcolor{blue}{http://www.ok.sc.e.titech.ac.jp/res/Stomach3D/}).










\bibliographystyle{IEEEtran}
\bibliography{Reference}

\end{document}